\documentclass[letterpaper, 10 pt, conference]{ieeeconf} 
\IEEEoverridecommandlockouts
\usepackage{cite}
\usepackage{amsmath,amssymb,amsfonts}
\usepackage{algorithmic}
\usepackage{graphicx}
\usepackage{textcomp}
\usepackage{xcolor}
\usepackage{siunitx}
\usepackage{balance}

\def\BibTeX{{\rm B\kern-.05em{\sc i\kern-.025em b}\kern-.08em
    T\kern-.1667em\lower.7ex\hbox{E}\kern-.125emX}}
\begin{document}

\title{\LARGE \bf Airflow Source Seeking on Small Quadrotors \\Using a Single Flow Sensor\\
}

\author{Lenworth Thomas$^{1}$, Tjaden Bridges$^{1}$, and Sarah Bergbreiter$^{1}$
\thanks{This research effort was supported by Air Force Office of Scientific Research MURI award number FA9550-19-1-0386. }
\thanks{$^{1}$ Mechanical Engineering, Carnegie Mellon University, Pittsburgh, PA, USA}
\thanks{Corresponding Author: Lenworth Thomas (lenwortt@andrew.cmu.edu)}
}



\maketitle

\begin{abstract}
As environmental disasters happen more frequently and severely, seeking the source of pollutants or harmful particulates using plume tracking becomes even more important. Plume tracking on small quadrotors would allow these systems to operate around humans and fly in more confined spaces, but can be challenging due to poor sensitivity and long response times from gas sensors that fit on small quadrotors. In this work, we present an approach to complement chemical plume tracking with airflow source-seeking behavior using a custom flow sensor that can sense both airflow magnitude and direction on small quadrotors ($<$~\SI{100}{\gram}). We use this sensor to implement a modified version of the `Cast and Surge' algorithm that takes advantage of flow direction sensing to find and navigate towards flow sources. A series of characterization experiments verified that the system can detect airflow while in flight and reorient the quadrotor toward the airflow. Several trials with random starting locations and orientations were used to show that our source-seeking algorithm can reliably find a flow source. This work aims to provide a foundation for future platforms that can use flow sensors in concert with other sensors to enable richer plume tracking data collection and source-seeking. 
\end{abstract}


\section{Introduction}

Understanding the sources and patterns of airflow can provide essential insights into the source and dispersion of gases or airborne particulates \cite{marino_evaluation_2015}. This can be helpful in identifying the source of gas leaks after natural disasters \cite{10160816}, environmental monitoring \cite{shaw_methods_2021}, and hazardous emissions from waste products \cite{BURGUES2022157290} among other applications. Given the hazard to humans in these applications, plume tracking has a rich history in the robotic space on ground-based vehicles \cite{groundplume}, underwater vehicles \cite{waterplume}, and unmanned aerial vehicles (UAVs) \cite{shigaki}. Plume tracking aims to detect, quantify, and, in some cases, follow the chemical or particle signals of interest to their source, a behavior that we call `source-seeking'. UAVs are particularly interesting because they can seek sources in 3D space and cover larger areas than may be feasible with fixed sensors \cite{marino_evaluation_2015}. 


Recently, many research efforts have focused on reducing the size of plume tracking and source-seeking UAVs. Smaller UAVs (e.g., $< $~\SI{100}{\gram}) can operate in more spatially constrained environments and are safer to operate near humans. Given the typical lower cost of these small robots, they can also be used in swarms that can cover larger areas with higher sensing resolution. There have been several examples of small quadrotors designed for source-seeking behaviors \cite{luo_flying_2018}\cite{shigaki}\cite{burg}\cite{castro}. These systems often use commercial gas sensors, custom-engineered flow sensors \cite{defay_customizable_2022}, or arrays of gas sensors to find gradients in the chemical signature. A significant challenge in these systems is that gas sensors respond slowly, resulting in long times to find a gradient and track it to its source. Gas sensors can also be significantly affected by rotor flow induced in UAVs \cite{ercolani_3d_2023}.

\begin{figure}[t]
    \centering
    \includegraphics[width=1\linewidth]{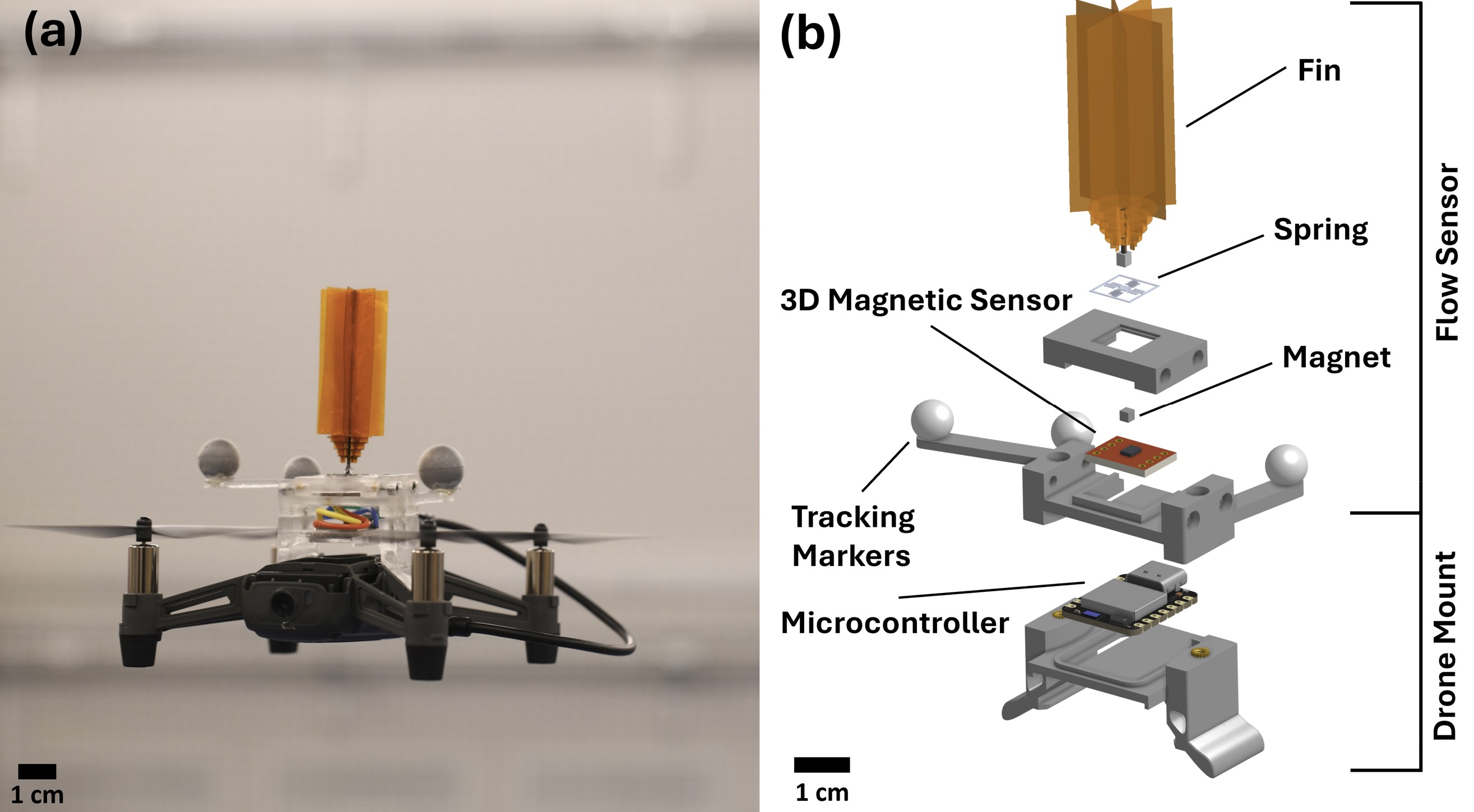}
    \caption{(a) A flying DJI Tello quadrotor with the flow sensor mounted on top. (b) Exploded view of the flow sensor and quadrotor mount. }
    \label{fig:quadrotorheadshot}
\end{figure}
Several recent approaches have been designed to overcome these challenges of detection speed and sensitivity for source-seeking behaviors on small quadrotors. Anderson et al. used a bio-hybrid approach by directly interfacing a moth antenna to the quadrotor as a chemical sensor \cite{anderson_bio-hybrid_2020}\cite{smellicopter}. The moth antenna enabled a fast response time but only has a limited set of detectable odors. In another approach, a swarm of small UAVs equipped with commercial gas sensors was used to find an alcohol source over a relatively larger area cluttered with obstacles \cite{sniffy}. This approach, Sniffy Bug, also took advantage of a novel gas dispersion model implemented on the small quadrotors \cite{sniffy}. Other approaches use novel algorithms to overcome slow and noisy sensors. For example, researchers have used reinforcement learning on small quadrotors for source-seeking tasks, although the source in this case was a light instead of a gas \cite{duisterhof2021learningseekautonomoussource}.

Another approach to overcoming the slow response of gas sensors on small UAVs for source-seeking tasks is to use measurement of the airflow itself to aid detection. This idea of anemotaxis (following a flow source) to complement chemotaxis (following a chemical source gradient) is hypothesized to be a primary mechanism in aiding insects in their exceptional odor-seeking skills \cite{Baker9383}\cite{BreugeulAnemo}. There are many methods by which to detect airflow on a UAV as reviewed in \cite{abimulti}, but many of these methods and sensors are only suitable for larger UAVs. Small UAVs can use pose estimation approaches that use IMU measurements combined with commanded rotor speeds, but these are often susceptible to significant drift over time \cite{wang_wind_2018}. Custom flow sensors can also be designed for small UAVs \cite{noauthor_whisker-inspired_nodate}\cite{deer_lightweight_2019}, but obtaining high quality flow data often requires multiple sensors spread over a large area, as seen in the large UAV in \cite{tagliabue_touch_2020}.


The primary contribution of this work is the integration of a single flow sensor and a bioinspired source-seeking algorithm on a small palm-sized quadrotor for airflow source-seeking (Fig. \ref{fig:quadrotorheadshot}). The flow sensor outputs both the magnitude and direction of airflow to enable rapid detection and response. After characterizing this sensor in flight tests, we demonstrate that we can quickly reorient the quadrotor toward the airflow source. The goal of this `flow servoing' approach is to ultimately aid future on-board (and often directional) gas or other environmental sensors by increasing their sensitivity and response time. These results are combined in demonstrations of a bio-inspired algorithm, Vector Surge, that is designed to find airflow in the environment and follow the airflow upstream to its source. This algorithm builds on previous bio-inspired algorithms like `Cast and Surge' \cite{smellicopter} by incorporating the airflow direction obtained by the flow sensor.

\section{Integrating a Flow Sensor With a Small Quadrotor}

\subsection{Flow Sensor Background}\label{AA}

The design of the flow sensor used in this project (Fig. \ref{fig:quadrotorheadshot}) was heavily inspired by the work of Kim et al. \cite{noauthor_whisker-inspired_nodate}. Airflow on the sensor fin creates a moment that rotates the fin and a magnet placed at the bottom of the fin with respect to a planar spring suspension. A magnetic field sensor is placed underneath the fin to measure the change in magnetic field that occurs from the displacement and rotation of the fin magnet in three axes, providing an estimate of the fin rotation. The magnitude of the rotation provides an estimate of the incident airflow velocity, while the direction of the fin movement provides an estimate of the incident airflow direction. This magnetic field measurement (in milliTesla) can be translated to rotation (in degrees) and airflow velocity (in meters/second) as seen in \cite{noauthor_whisker-inspired_nodate}, but for this work, we have focused on the rotation estimated from the $B_x$ and $B_y$  components. We have left the magnitude response of the airflow in units of milliTesla. It is important to note that the current flow sensor design provides 2D flow vectors; 3D source-seeking is left for future work.

\subsection{Quadrotor Integration}
These flow sensors have previously been integrated on larger quadrotors in \cite{tagliabue_touch_2020}, but the goal of this work is a lightweight integration with a small quadrotor that can access more spatially constrained environments and operate around humans. Two aspects of this integration need to be addressed: physical mechatronic integration with the quadrotor and signal processing from the flow sensor. The quadrotor selected for this work was the DJI Tello. With a footprint of \SI{98}{\milli\meter} x \SI{92.5}{\milli\meter} and a weight of \SI{80}{\gram}, the Tello is one of the smallest programmable quadrotors available. A significant benefit of the Tello over other small quadrotors is that it can support an \SI{80}{\gram} payload, making it easier to test new hardware quickly with minimal change in flight performance.  In addition, the programming interface made high-level position control simple. However, the approach of airflow sensing with a single flow sensor applies to any small-scale quadrotor.

\subsubsection{Mechatronic Integration} To make the integration of the flow sensor as modular as possible, a separate microcontroller is used to capture sensor data. This microcontroller can either interface with the quadrotor's microntroller for autonomous operation or can be interfaced wirelessly to send data to a computer for further analysis. In this case, a Seeed Studio XIAO nRF52840 board was used to interface with the sensor and send data over Bluetooth Low Energy (BLE) to a laptop running a Python script that controls the quadrotor over WiFi. The microcontroller is powered directly by the quadrotor's USB port which provides a \SI{5}{\volt} power supply. 

The flow sensor was mounted to the quadrotor via a 3D-printed mount (Fig. \ref{fig:quadrotorheadshot}). This mount was also designed to be modular so that different fin configurations could be evaluated. The sensor and mount weigh a combined \SI{15.6}{\gram} resulting in a total mass $< $~\SI{100}{\gram}. The height of the fin placement on the mount was informed by previous work showing that planar flow from the quadrotor's rotors is minimal at a location above the center of the quadrotor \cite{carter_influence_2020}. When the quadrotor is hovering, the disturbance from vibrations and turbulence is minimal. Even when hovering, the system can detect airflow at speeds greater than or equal to \SI{0.2}{\meter\per\second}, which is designated as `Calm' on the Beaufort Wind Scale \cite{National_Weather_Service_2016}, a standardized metric for measuring wind intensity.  

\subsubsection{Flow Signal Processing} 
\begin{figure}
    \centering
    \includegraphics[width=1\linewidth]{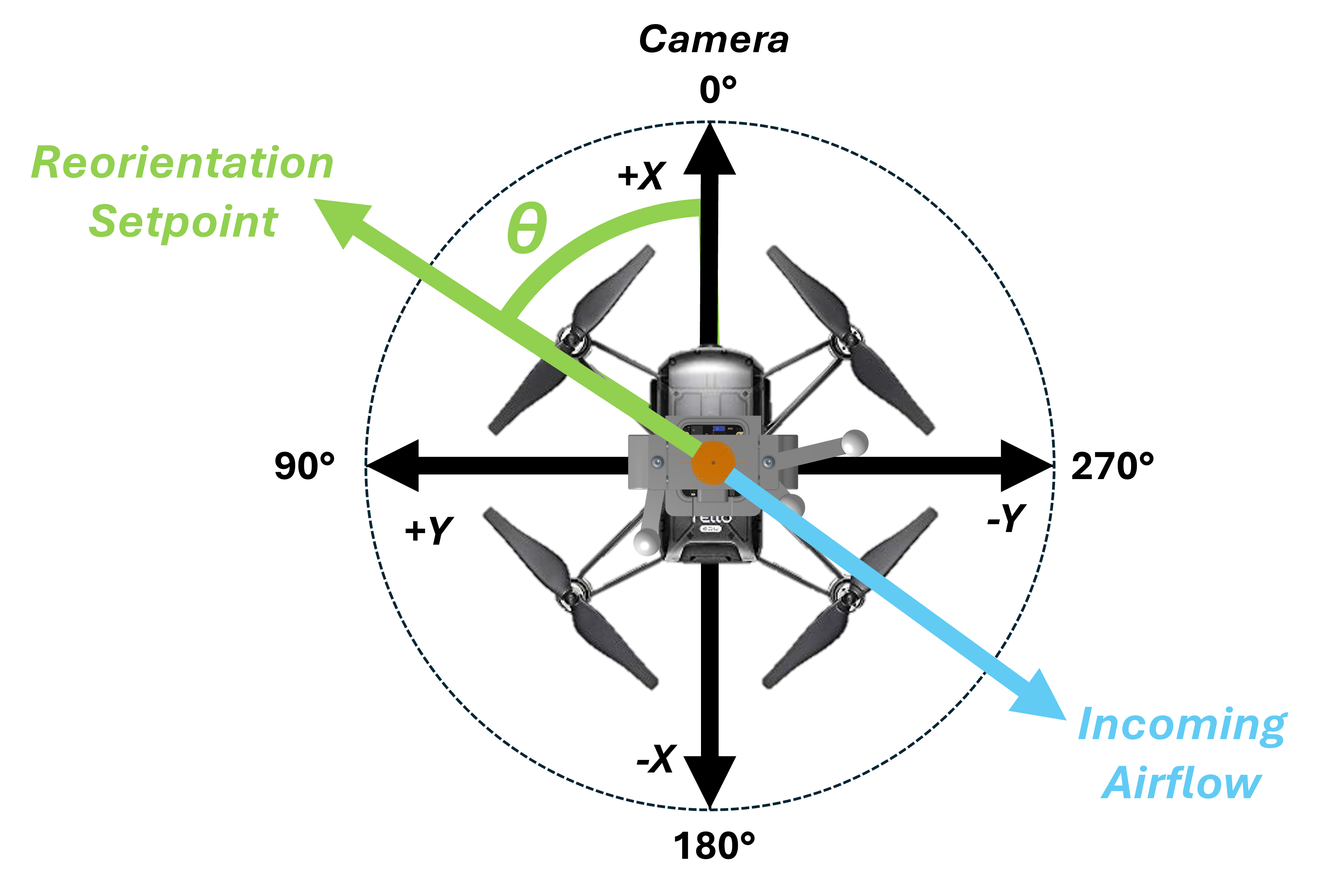}
    \caption{The quadrotor coordinate system and a representation of the angle, $\theta$, that defines the error between the quadrotor heading (\SI{0}{\degree}) and the upwind direction.}
    \label{fig:coord}
\end{figure}

The microcontroller samples the $B_x$ and $B_y$ readings from the flow sensor at \SI{40}{\hertz}. A calibration process is used at take-off to account for any offsets from sensor manufacturing and effectively `zero' the sensor. To minimize sensor noise, a moving average filter was used to filter the $B_x$ and $B_y$ signals. Unless otherwise noted, a 10-point moving average filter was used. The resulting $B_x$ and $B_y$ signals are then transformed to quadrotor coordinates $X_{quadrotor}$ and $Y_{quadrotor}$. To calculate the angle of airflow with respect to the quadrotor, we can use the inverse tangent of ($Y_{quadrotor}/X_{quadrotor}$). This angle provides the direction in which the fin has moved. To calculate the direction of an airflow source, $\theta$ (as shown in Fig. \ref{fig:coord}), we need to add an additional \SI{180}{\degree}. The direction of the incoming airflow in degrees is expressed as follows.

\begin{equation}
\theta = \arctan\left(\frac{Y_{\text{quadrotor}}}{X_{\text{quadrotor}}}\right) \cdot \frac{\SI{180}{\degree}}{\pi} + \SI{180}{\degree}
\end{equation}

Magnitude of flow is calculated as $\lvert B \rvert = \sqrt{B_x^2 + B_y^2}$.



\section{Methods}
The goal of this work is to demonstrate airflow source-seeking on a small quadrotor. To that end, the first set of experiments characterize the flow sensor's ability to detect the incident angle of flows with different magnitudes while in flight subject to rotor flow and vibrations. The second set of experiments explore the quadrotor's ability to orient toward a flow using data directly from the flow sensor. The final set of experiments implement the Vector Surge algorithm that uses the measured flow vector to both orient a sensor (e.g., gas sensor) towards a flow and move the quadrotor upwind. 

\subsection{Flow Sensor Characterization in Flight}

\begin{figure}[t]
    \centering
    \includegraphics[width=1\linewidth]{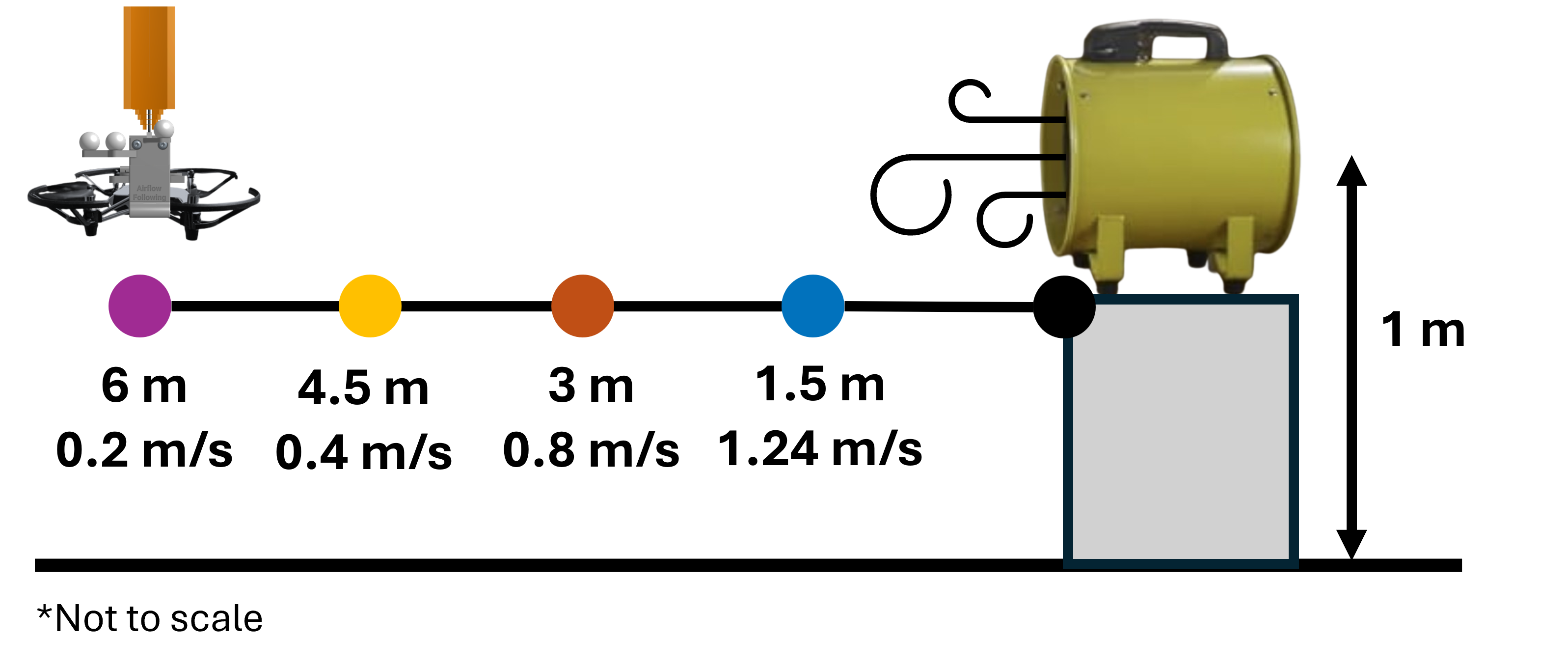}
    \caption{The four measurement locations and their respective distances and average airflow speed measured by an anemometer. The fan was fixed at a height of 1 m off the ground, level with the quadrotor's height in the air. }
    \label{fig:flowSetup}
\end{figure}

Flow sensor characterization and calibration using a wind tunnel was previously demonstrated in \cite{noauthor_whisker-inspired_nodate}. 
In this work, we characterize the ability of the flow sensor to measure the angle of incident airflow while hovering, and the experiment is illustrated in Fig. \ref{fig:flowSetup}. A fan (Vevor Utility Fan with a flow rate of \SI{1948} CFM) was placed \SI{1}{\meter} above the ground, level with the quadrotor's takeoff height. The quadrotor was positioned at four different distances relative to the fan (\SI{1.5}{\meter}, \SI{3}{\meter}, \SI{4.5}{\meter}, and \SI{6}{\meter}) so that the quadrotor was subject to four different average airflow speeds (\SI{1.24}{\meter\per\second}, \SI{0.8}{\meter\per\second}, \SI{0.4}{\meter\per\second}, and \SI{0.2}{\meter\per\second} respectively). These speeds are all within the Calm and Light Air categories of the Beaufort Wind Scale (Beaufort Numbers 0 and 1 respectively) \cite{National_Weather_Service_2016}. 
A handheld anemometer (Thomas Scientific 1235D16) was used to measure the average airflow speed at these locations. In theory, the maximum flow speed the system can handle depends on the quadrotor itself. Airflows over \SI{1.5}{\meter\per\second} typically pushed the quadrotor so experiments were conducted under this speed.

This experiment was run in a \SI{10}{\meter} x  \SI{10}{\meter} indoor motion capture arena using 24 Optitrack PrimeX 41 cameras and the Optitrack Motive software to provide a ground truth of the quadrotor's positioning. At all four locations, the quadrotor was commanded to take off, hover for \SI{3}{\second}, and then run the sensor calibration routine. At this point, the fan was manually turned on. After a waiting period of \SI{10}{\second} to ensure that the fan reaches its maximum speed, the quadrotor was commanded to rotate in place around its yaw axis for \SI{60}{\second} (achieving approximately four full rotations during this time). During these \SI{60}{\second}, the off-board computer logs the quadrotor's yaw angle as calculated from its IMU and the flow sensor's magnetic field components ($B_x$ and $B_y$) to calculate airflow angle, $\theta$. Motion tracking data were also recorded to track the quadrotor position and orientation with respect to the fan.

\subsection{Reorienting Towards Flow}

Many of the sensors that we would like to incorporate for flow source detection such as the eNose board that can measure volatile organic compounds\cite{Rossi} or sensors that measure PM2.5 particulates\cite{piotr-pollution}, are directional in nature; they have the highest sensitivity at a specific orientation. 
The following experiment aims to demonstrate that we can use the quadrotor itself to reorient a sensor for improved readings by always directing the sensor toward airflow. This capability provides a building block for the Vector Surge algorithm. 



In this experiment, the quadrotor is commanded to take off, hover, calibrate, and then the fan is manually turned on similar to the previous experiment. The system waits for \SI{10}{\second} for the fan to reach maximum speed. After these \SI{10}{\second}, the reorientation controller is turned on. Using only the measured airflow angle, $\theta$, from the flow sensor, the system uses a PD controller with gains of $K_p = $~\num{0.6} and $K_p = $~\num{0.08} to control the quadrotor's yaw rate. The quadrotor turns to minimize the angle of the incoming airflow until the airflow aligns with the front of the quadrotor. The experiment was repeated at the same four locations and airflows as shown in Fig. \ref{fig:flowSetup} for three different starting orientations: \SI{180}{\degree}, \SI{90}{\degree}, and \SI{45}{\degree} relative to the airflow. The goal of this experiment is to reorient within the time of a typical to a sustained gust as defined by the American Meterological Society, \SI{3}{\second} to \SI{20}{\second}  \cite{Gust}.

\subsection{In-Plane Source-Seeking}
\begin{figure}
    \centering
    \includegraphics[width=1\linewidth]{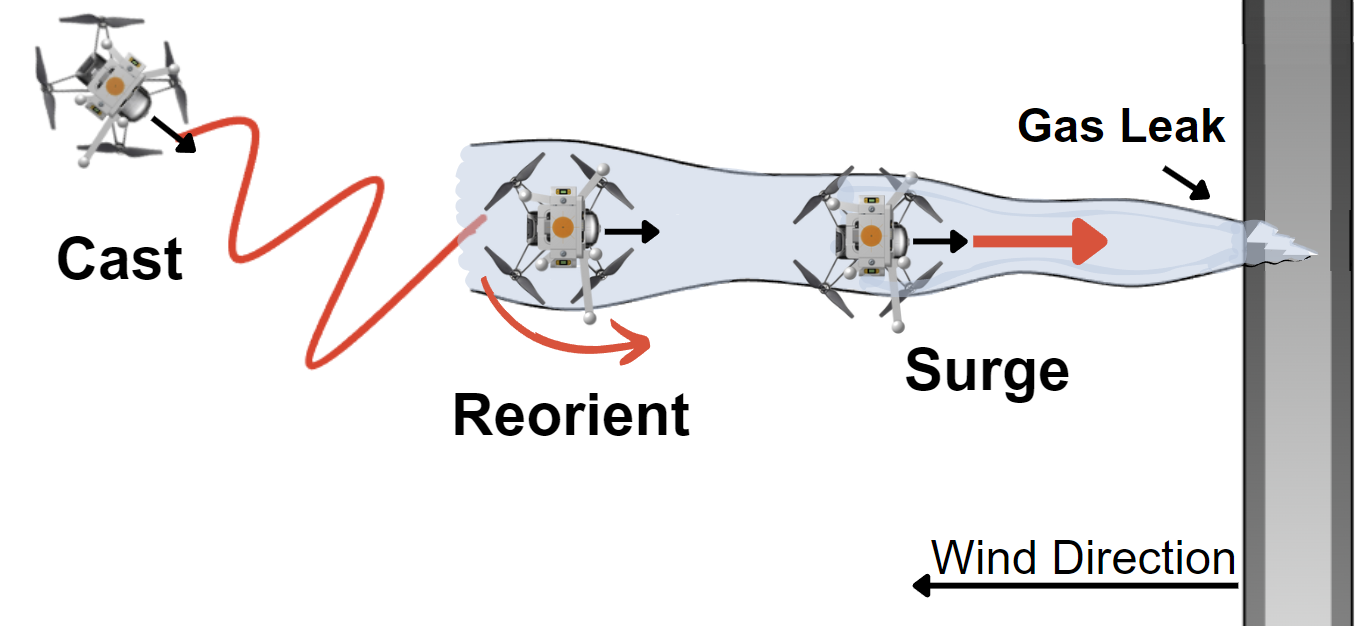}
    \caption{Behavior of the quadrotor using the Vector Surge algorithm.}
    \label{fig:castandflow}
\end{figure}

In some applications, such as a gas leak, the quadrotor may have to find the airflow before seeking the source. This experiment demonstrates our system's ability to discover in-plane flows (Fig. \ref{fig:castandflow}) with the on-board flow sensor by adapting a bio-inspired `cast and surge' algorithm frequently used for plume tracking (e.g., \cite{anderson_bio-hybrid_2020}). This algorithm is derived from the behavior of a moth that `casts' back and forth until detecting a scent of interest at which point the moth surges upwind in hope of finding the source of the scent.

\begin{figure}
    \centering
    \includegraphics[width=0.75\linewidth]{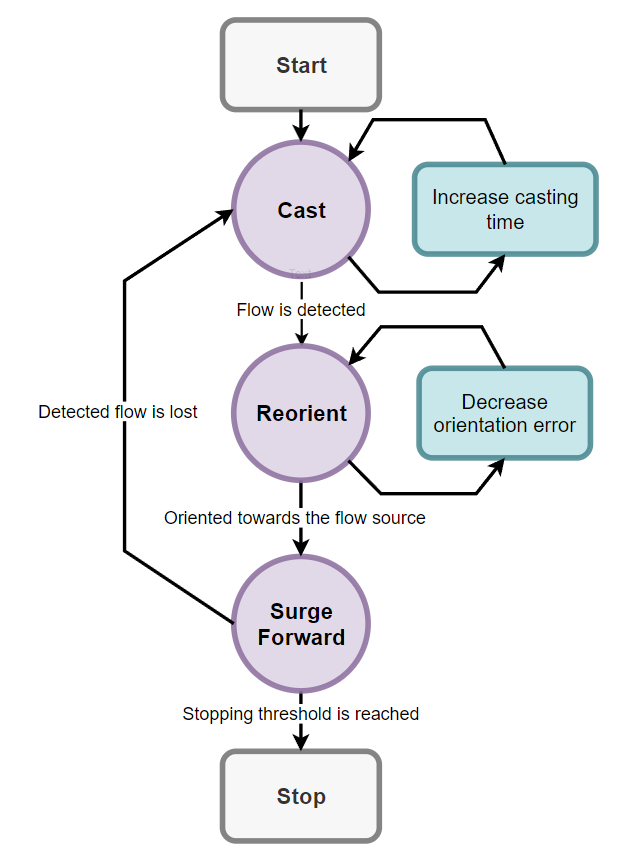}
    \caption{Flowchart depicting each state in Vector Surge and each respective operation.}
    \label{fig:castFSM}
\end{figure}

We implement the Vector Surge algorithm described in Fig. \ref{fig:castFSM} to take advantage of the flow vector (including both magnitude and direction) measured by the on-board flow sensor.
After takeoff and initial quadrotor calibration, the cast state is enabled if no flow is detected. In the cast state, the system performs a set of translational movements; moving left, forward, and then right. Each time the quadrotor repeats these movements, the commanded movement time increases with every cast to effectively widen the area being searched. 

The algorithm uses a defined flow magnitude threshold to confirm that the flow has been detected. As highlighted in previous experiments, the flow sensor is responsive to flow generated from the quadrotor's motion as well as the rotors in hover. The detection threshold was set at \SI{4.6}{\milli\tesla}, \SI{120}{\percent} of the maximum recorded flow magnitude during casting (with a maximum commanded quadrotor velocity of \SI{0.2}{\meter\per\second}). 
In the future, this threshold can be reduced through improved state estimation with the IMU and flow sensor. 


        
Once flow is detected, the system enters the reorient state. This state uses the PD controller discussed previously to rotate the quadrotor so that a sensor (in this case, the quadrotor's camera) is oriented toward the flow source. Once reorientation is complete, the algorithm moves to the surge state in which the quadrotor will move forward towards the airflow. In this state, the quadrotor continually checks the flow signal to determine if a maximum flow magnitude has been reached (the stop condition). If the flow source is lost during a surge, the system returns to the cast state.

Experiments using the Vector Surge algorithm were run in the same motion capture arena to track both the quadrotor's position and orientation over the course of the algorithm. The flow signal was also recorded during the experiment. This experiment was run ten times with randomized quadrotor starting locations and orientations with respect to the fan.

\section{Results}
\subsection{Flow Sensor Characterization in Flight}



\begin{figure}
    \centering
    \includegraphics[width=1\linewidth]{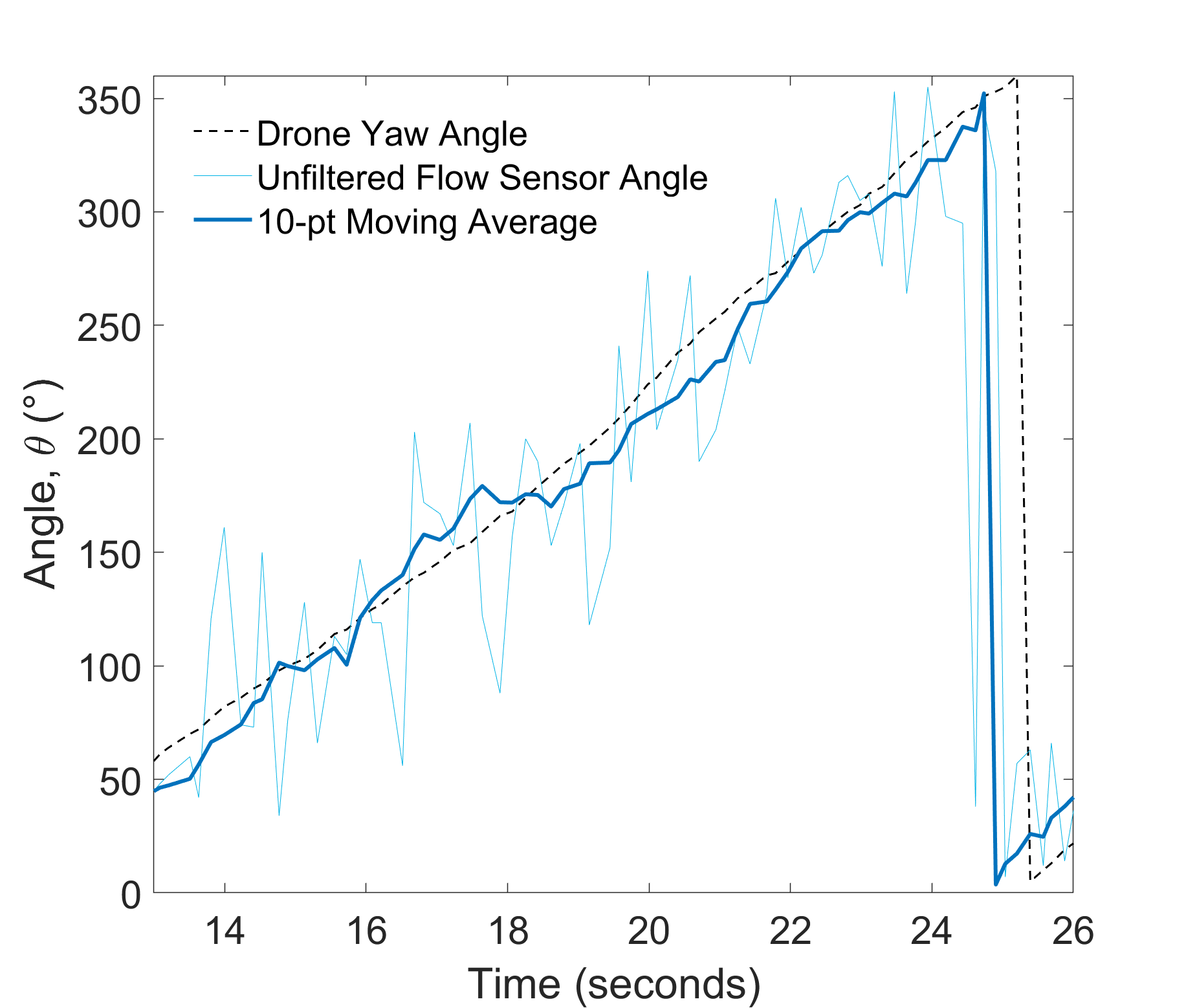}
    \caption{Filtered and unfiltered angle estimation from the flow sensor and the quadrotor's IMU yaw angle at \SI{1.24}{\meter\per\second}. A 10-point moving average was chosen due to its combination of speed and stability.}
    \label{fig:Sawtooth}
\end{figure}

\begin{figure}[t]
    \centering
    \includegraphics[width=1\linewidth]{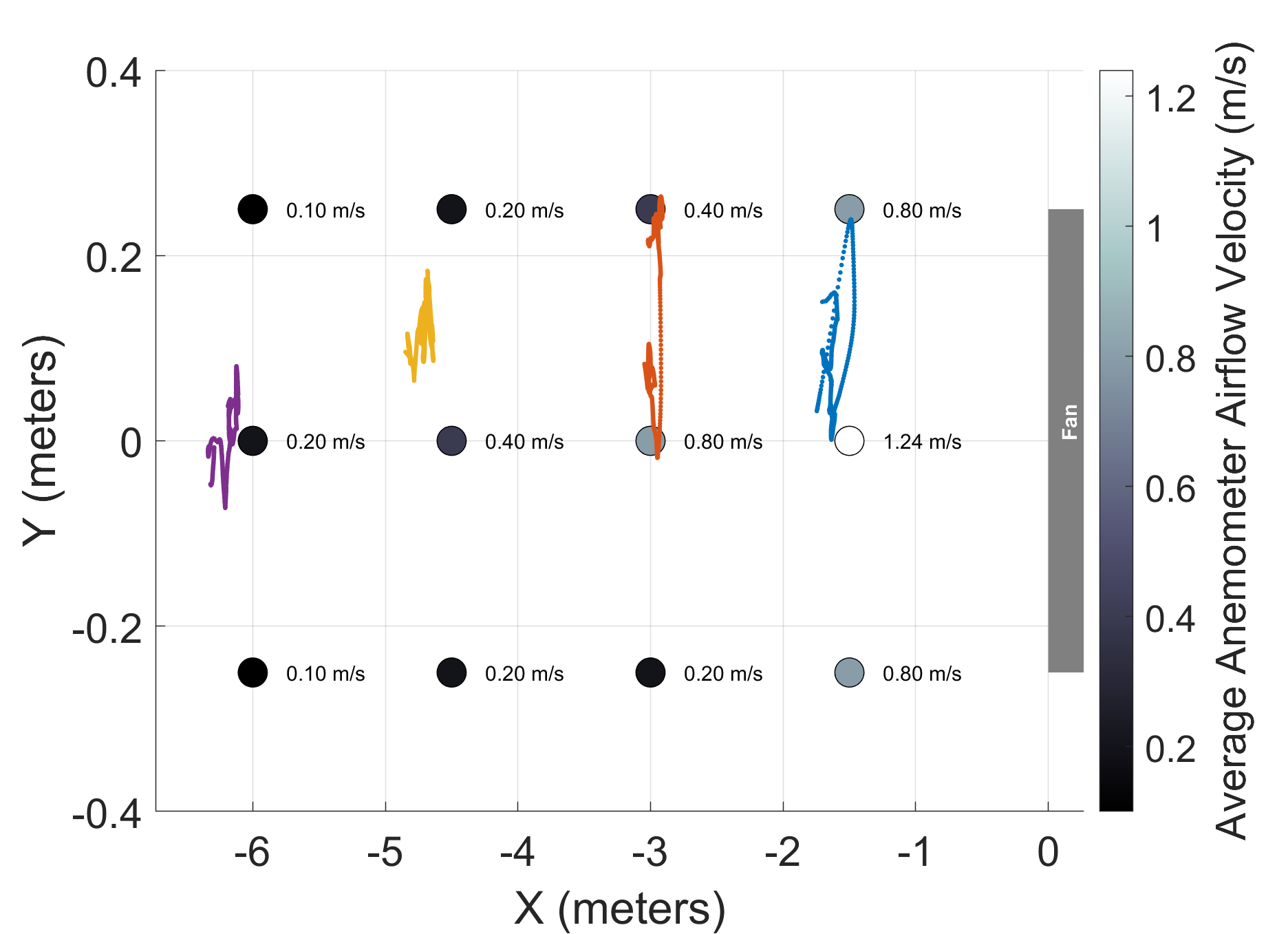}
    \caption{The translation of the quadrotor over a \SI{60}{\second} experiment under flow at different distances away from the fan. The anemometer readings were averaged at each point over \SI{10}{\second}.}
    \label{fig:TranslationDuringFlight}
\end{figure}

In this experiment, we were interested in measuring the airflow angle relative to the quadrotor as the quadrotor rotates around its yaw axis. Fig. \ref{fig:Sawtooth} shows a yaw angle estimate calculated from the quadrotor's IMU, the raw flow sensor readings, and the filtered signal from a 10-point moving average (modulo \SI{360}{\degree} due to the continuous rotation). The raw flow sensor data is noisy, but the moving average tracks the yaw angle relatively closely providing a good combination of response time and stability. 

To further investigate the source of noise in the flow sensor measurements, we took a closer look at the motion of the quadrotor throughout the course of these rotation tests as captured by the motion tracking system. Fig. \ref{fig:TranslationDuringFlight} shows the quadrotor motion throughout the tests at four different positions, resulting in four different airflow velocities. Markers in the background show average airflow velocities as measured by the anemometer at these locations to provide a sense of how airflow might change as the quadrotor  moves in and out of the airflow over the course of the experiment. In higher airflow, the small quadrotor struggles to maintain its position and the airflow speed incident on the quadrotor over the course of the experiment varies from \SI{1.24}{\meter\per\second} to \SI{0.8}{\meter\per\second} at the position closest to the fan.

\begin{figure}[t]
    \centering
    \includegraphics[width=1\linewidth]{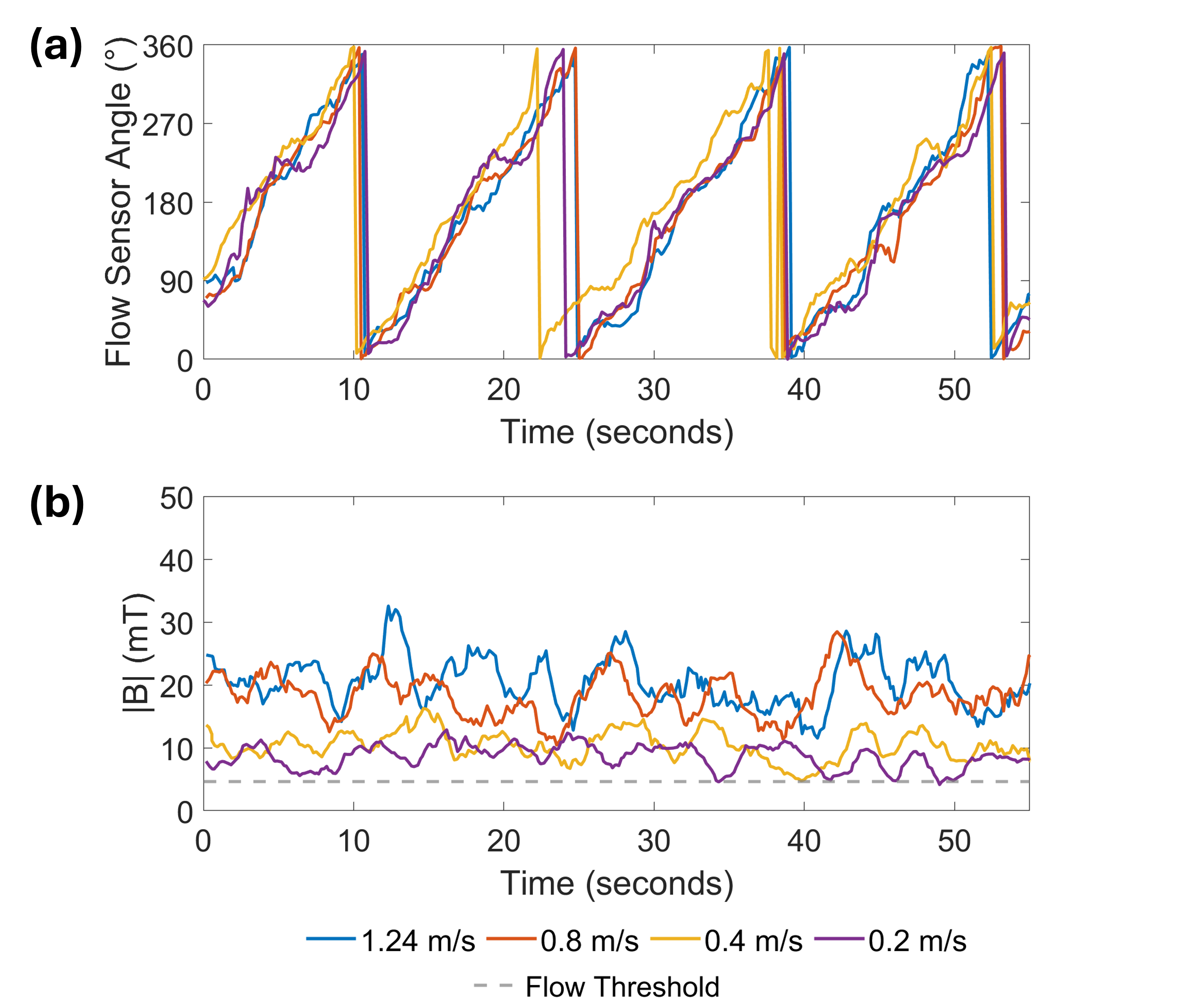}
    \caption{(a) Filtered flow sensor angle measurements across all four airflow velocities as the quadrotor rotates continuously from \SIrange{0}{360}{\degree} over \SI{60}{\second}.  (b) Magnitude measurements at four airflow velocities during rotation. }
    \label{fig:Characterization}
\end{figure}

\begin{figure}[tb]
    \centering
    \includegraphics[width=1\linewidth]{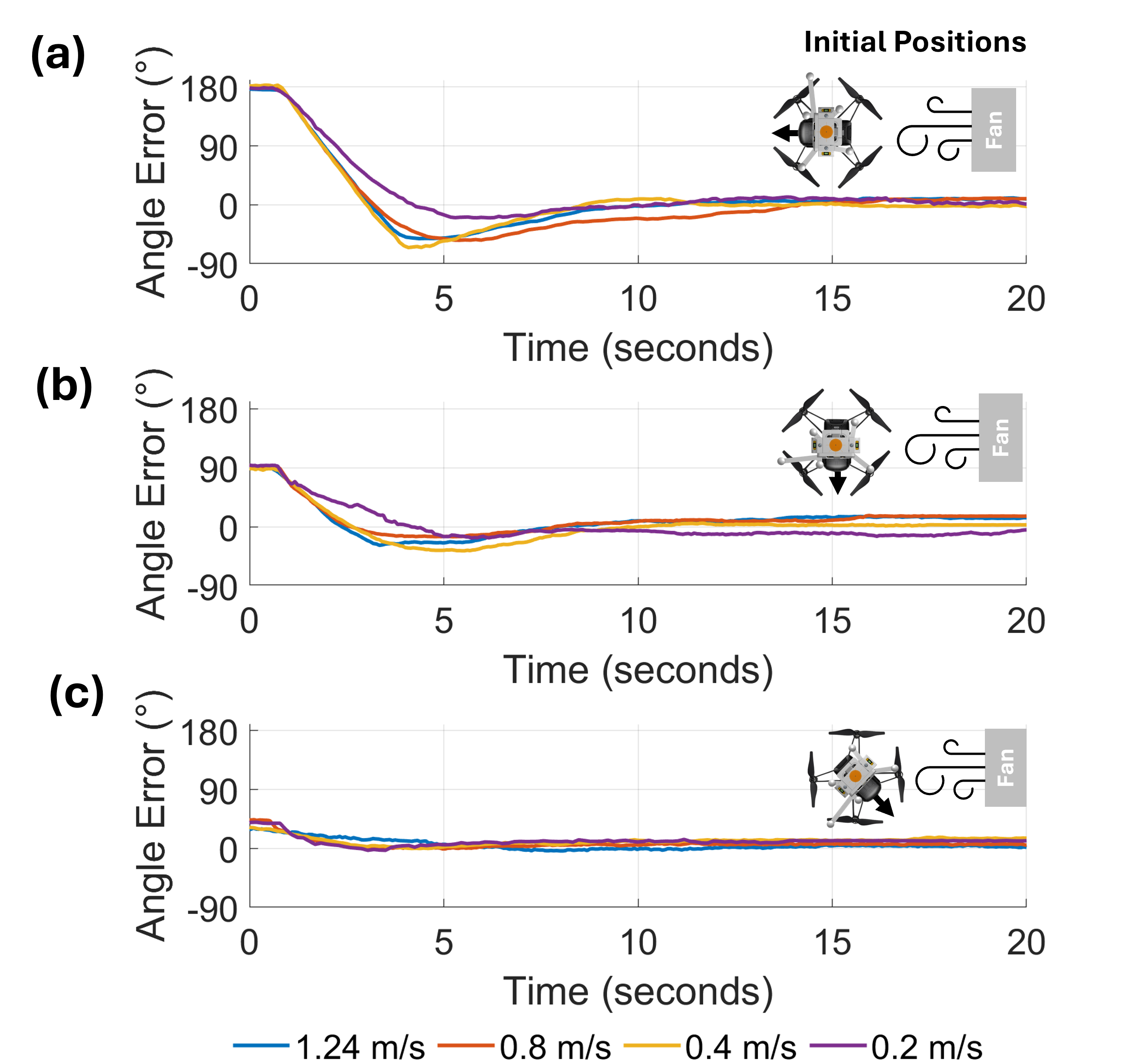}
    \caption{a-c) Angle error from reorientation experiments starting at $\theta = $~\SI{180}{\degree}, \SI{90}{\degree}, and \SI{45}{\degree} respectively.}
    \label{fig:OrientTest}
\end{figure}

With these results in mind, Fig. \ref{fig:Characterization} shows the results of continuous quadrotor rotation with respect to airflow at these four different locations and four different average airflow velocities. The calculated angle of airflow closely follows the yaw rotation angle provided by the IMU at all four airflow velocities as shown in Fig. \ref{fig:Characterization}a. As expected, in Fig. \ref{fig:Characterization}b the average magnitude measured by the flow sensor decreases as the velocities decrease farther away from the fan. The motion of the quadrotor in the flow is especially apparent in these data. Despite this effect on magnitude, in-flight flow sensor characterization confirmed that the flow sensor can reliably estimate the angle of incident flow relative to the quadrotor over a wide range of airflow velocities, \SI{0.2}{\meter\per\second} to \SI{1.2}{\meter\per\second}.  While noisy, the magnitude signal is reliable enough to detect airflow above the threshold. 

\begin{figure*}
    \centering
    \includegraphics[width=1\linewidth]{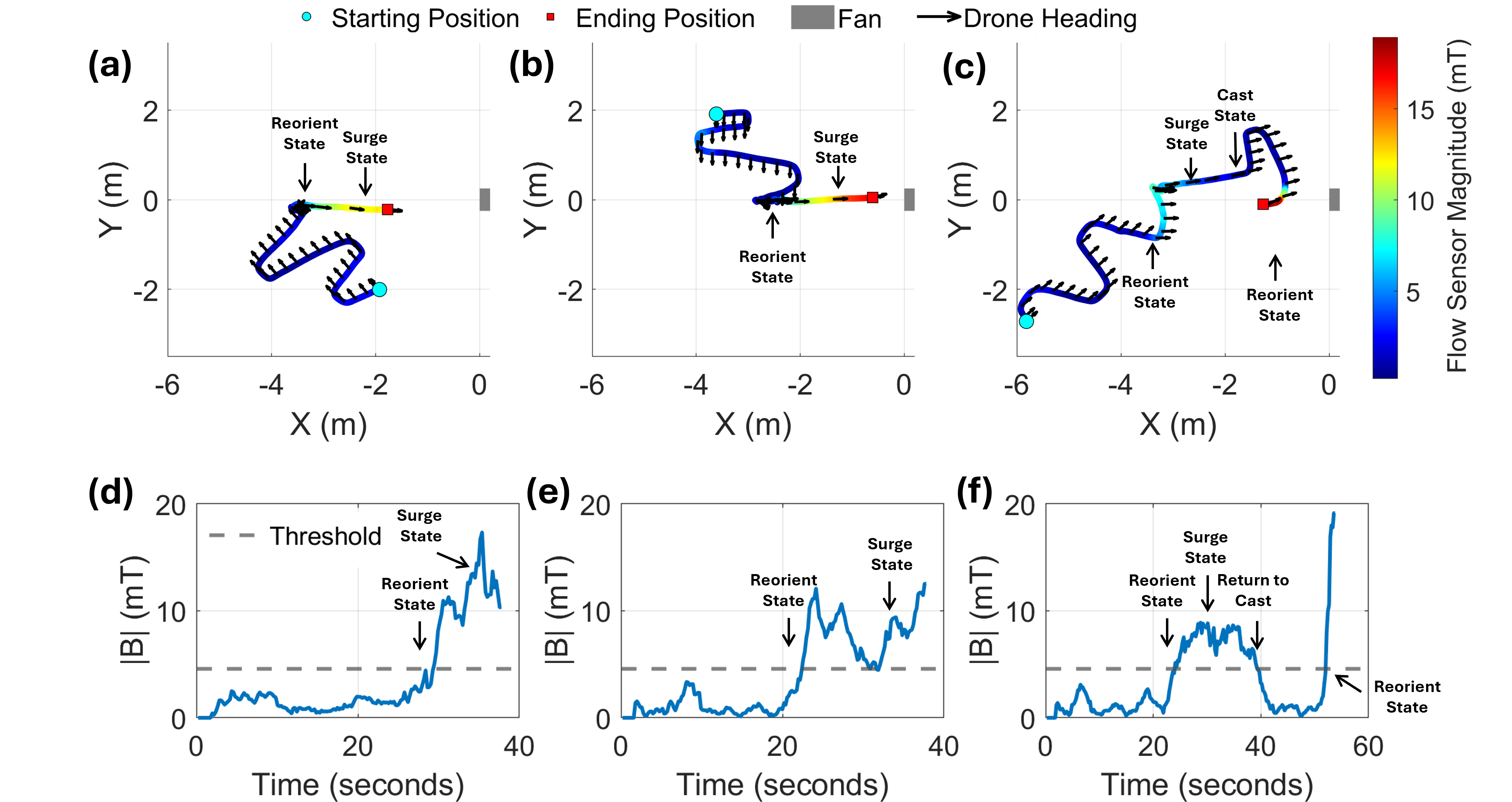}
    \caption{Three sample trials of the quadrotor executing the Vector Surge algorithm with different starting positions and orientations in the testing arena. (a-c) Each X-Y trajectory with arrows representing quadrotor orientations is plotted above (d-f) its respective flow magnitude plot.}
    \label{fig:Trajectories}
\end{figure*}


\subsection{Reorienting Towards Flow}

Fig. \ref{fig:OrientTest} shows the angle orientation error versus time results for the reorientation tests. The error shown in the figure is the difference between the quadrotor's yaw angle and the airflow as measured by the OptiTrack system. Performance was relatively similar across the range of airflows tested, and the quadrotor was typically able to reorient to within \SI{20}{\degree} of the airflow direction in \SI{10}{\second}, approximately half the time of a typical sustained gust \cite{Gust}. In trials that did not settle to a \SI{0}{\degree} error, the quadrotor was typically drifting out of the strongest airflow leading to a less accurate angle estimate from the flow sensor. 
 
\subsection{In-Plane Source-Seeking}

Three representative trials of the Vector Surge algorithm are shown in Fig. \ref{fig:Trajectories}. This figure shows both the motion of the quadrotor in the X-Y plane as well as the orientation of the quadrotor represented by arrows through its trajectory. The airflow magnitude from the sensor is represented as a heatmap on the trajectory as well as in the respective figures below each trajectory. For example, in the first trial (Fig. \ref{fig:Trajectories}a), the quadrotor started oriented and moving away from the fan. When flow was detected approximately \SI{3.5}{\meter} away from the fan at $t = $~\SI{29}{\second}, the quadrotor reoriented and surged toward the fan before reaching its stop condition. In the trial represented in Fig. \ref{fig:Trajectories}b, the quadrotor is blown backwards briefly while reorienting, but it is able to recover and surge to the airflow source. In Fig. \ref{fig:Trajectories}c, the quadrotor was able to sense flow, reorient, and surge, but moved out of the airflow during the surge at $t = $~\SI{40}{\second}. As a result, it returned to the cast state at which point it was able to quickly find the airflow again and reach its stop condition. The three trials in Fig. \ref{fig:Trajectories} completed in \SIrange{38}{55}{\second}.


In a few cases, the quadrotor failed to find the source. Trials were stopped manually before the quadrotor might hit the walls of the motion capture arena before airflow was detected. This could be avoided in future implementations by including obstacle avoidance similar to \cite{sniffy}. On at least one occasion, the quadrotor was unable to detect airflow during casting because the flow at all locations traversed by the quadrotor was below our \SI{0.2}{\meter\per\second} threshold. Adding improved state estimation on-board the quadrotor could correct for quadrotor motion in the flow sensing signal to enable detection of even lower airflow velocities. 

\section{Conclusions}
In this work, we demonstrated the use of a custom flow sensor and a new algorithm to seek airflow sources on small quadrotors. The results show that we are able to reliably detect airflow direction at flow velocities as low as \SI{0.2}{\meter\per\second}. We also show that we can reorient the quadrotor toward the airflow which can ultimately help improve sensitivity and response time of chemical sensors on the quadrotor. Finally, we showed that starting from a random location and orientation with respect to an airflow source, we could find the airflow and navigate to the source quickly and repeatedly. 

While the system responded well under the relatively planar and controlled conditions in the experiments in this work, there are several limitations to our current approach. First, flow acting on the sensor from above or below may still be detected, but not reliably; thus this approach is currently limited to flow in-plane with our current sensor design. Additionally, to make this work more useful in complex environments, we would need obstacle detection algorithms and state estimation algorithms to help separate flow on the sensor from flight versus flow from an air source. Finally, using this approach with chemical sensing would allow us to follow flow sources only when a chemical of interest was also present making the algorithm more suitable for outdoor environments. 

\balance

\bibliographystyle{ieeetr}
\bibliography{bibliography}
\vspace{12pt}

\end{document}